\documentclass[journal,10pt]{IEEEtran}

\usepackage{hyperref}
\usepackage{verbatim}
\usepackage{graphicx}
\usepackage{multicol}
\usepackage{amsmath}
\usepackage{amssymb}  % 提供\checkmark命令
\usepackage{cuted}
\usepackage{cite}
\usepackage{multirow}
\usepackage{hhline}
\usepackage{xcolor}
\usepackage{balance}
\usepackage{subfigure}
\usepackage{hyperref}
\usepackage{pifont}
\usepackage{booktabs} 
\usepackage{stfloats}
\usepackage{makecell}

\makeatletter

\def\hlinewd#1{%
	\noalign{\ifnum0=`}\fi\hrule \@height #1 \futurelet
	\reserved@a\@xhline}

\makeatother
	
\begin{document}
	
\title{TCFormer: A 5M-Parameter Transformer with Density-Guided Aggregation for Weakly-Supervised Crowd Counting}

\author{Qiang Guo, Rubo Zhang, Bingbing Zhang, Junjie Liu and Jianqing Liu
	
	\thanks{Qiang Guo is with College of Mechanical and Electronic Engineering, Dalian Minzu University, 116650, Dalian, China, and also with Dalian University of Technology and Postdoctoral workstation of Dalian Rijia Electronics Co., Ltd., 116630, Dalian, China. 
		(e-mail: guoqiang01486@dlnu.edu.cn).}% <-this % stops a space
	\thanks{Rubo Zhang and Junjie Liu are with College of Mechanical and Electronic Engineering, Dalian Minzu University, 116650, Dalian, China. (e-mail: zhangrubo@dlnu.edu.cn, junjie1125@dlnu.edu.cn).}
	%\thanks{Manuscript received XXX, XX, 2015; revised XXX, XX, 2015.}
	\thanks{Bingbing Zhang is with School of Computer Science and Engineering, Dalian Minzu University, 116650, Dalian, China.
		(e-mail: icyzhang@dlnu.edu.cn).}
	\thanks{Jianqing Liu is with the R\&D Department of Dalian Rijia Electronics Co., Ltd., 116630, Dalian, China. (e-mail: liujq0806@163.com).}
	\thanks{\emph{Corresponding author: Qiang Guo}.}
}

\maketitle

\begin{abstract}
Crowd counting typically relies on labor-intensive point-level annotations and computationally intensive backbones, restricting its scalability and deployment in resource-constrained environments. To address these challenges, this paper proposes the TCFormer, a tiny, ultra-lightweight, weakly-supervised transformer-based crowd counting framework with only 5 million parameters that achieves competitive performance. Firstly, a powerful yet efficient vision transformer is adopted as the feature extractor, the global context-aware capabilities of which provides semantic meaningful crowd features with a minimal memory footprint. Secondly, to compensate for the lack of spatial supervision, we design a feature aggregation mechanism termed the Learnable Density-Weighted Averaging module. This module dynamically re-weights local tokens according to predicted density scores, enabling the network to adaptively modulate regional features based on their specific density characteristics without the need for additional annotations. Furthermore, this paper introduces a density-level classification loss, which discretizes crowd density into distinct grades, thereby regularizing the training process and enhancing the model's classification power across varying levels of crowd density. Therefore, although TCformer is trained under a weakly-supervised paradigm utilizing only image-level global counts, the joint optimization of count and density-level losses enables the framework to achieve high estimation accuracy. Extensive experiments on four benchmarks including ShanghaiTech A/B, UCF-QNRF, and NWPU datasets demonstrate that our approach strikes a superior trade-off between parameter efficiency and counting accuracy, outperforming many lightweight and heavy, fully-supervised counters while $>20\times$ fewer parameters and no location annotations. Thus, it can be a good solution for crowd counting tasks in edge devices. 
\end{abstract}

%The embedded deployment experiment demonstrates that the established pedestrian detection model can achieve a better comprehensive prediction ability by adding the proposed algorithms in the different stages according to the task demands.
%this work is the use of cascading strategy and learned intrinsic features to enhance the model's false positive suppression capability for pedestrian detection in its training and testing stages.
%In the testing stage, Intrinstic Refined Prpopsal algorithm and Intrinstic Refined Prpopsal plus Cascade Refined Prpopsal algorithms are adopted respectively to steadily improve the model's ability to remove false positives in its inference stage.
% List 3 to 10  pertinent keywords specific to the article, yet reasonably common within the subject discipline.
\section{Introduction}\label{sec1}

Crowd counting, the task of estimating the number of individuals in crowded scenes, is a fundamental problem in computer vision with critical applications in video surveillance, public safety, and traffic control,. etc. Given its importance across diverse domains, both academia and industry engineers have devoted substantial research efforts over the past decades. The remarkable progress achieved through evolving methodologies, from traditional image processing to modern deep learning, highlights the notable advancements in algorithmic paradigms, particularly with the emergency of convolutional neural networks (CNNs)~\cite{zhang2016single, li2018csrnet} and the recent rise of Vision Transformers (ViTs)~\cite{liang2022, lin2022} have further pushed the state-of-the-art prediction accuracy in crowd counting. However, although gold standards in this research field are stilling refreshed, the two persistent challenges: annotation cost and computational complexity hinder their widespread and practical deployment, which can be contributed into the following two problems. 

Firstly, the dominant fully-supervised paradigm relies on dense, pixel-level supervision. Annotating every head in thousands of crowded images with point annotations is extremely labour-intensive, error-prone and often impractical for large-scale, real-world datasets. This has led to a growing interest in weakly-supervised approaches that adopt weaker forms of supervision, e.g. global head counts, as discussed in {lei2021, wang2023}. However, achieving high predictive accuracy under such weak supervision remains challenging. Existing approaches that rely solely on global head counts from images are limited. The lack of precise spatial annotations — particularly the positional coordinates of individuals within images — makes it difficult to accurately describe the distribution of crowds, which is a particular issue when predicting the number of people in highly crowded images.

Secondly, conventional convolutional neural networks are constrained by their inherent inductive biases, while standard Vision Transformers face the well-known bottleneck of quadratic computational complexity relative to image size. Consequently, existing methods either have insufficient prediction accuracy, which renders them impractical for precise applications, or they contain tens of millions of parameters, which makes them too computationally intensive for resource-constrained environments such as edge devices. Therefore, achieving an optimal balance between prediction performance and model size remains a critical and unresolved challenge in network architecture design. Current lightweight CNN- and Transformer-based paradigms for crowd counting have yet to adequately address this fundamental trade-off.

%state-of-the-art models rely on heavy backbones that contain tens of millions of parameters,  making them computationally prohibitive for resource-constrained environments like edge devices. While Transformers excel at capturing global contextual information, their quadratic computational complexity relative to image size is a well-known bottleneck. 

%most high-performance counters—whether fully or weakly supervised—rely on heavy backbones (e.g., VGG-16, ResNet-50, or ViT-Base) containing tens of millions of parameters. While Transformers excel at modeling long-range dependencies, their quadratic computational complexity poses a severe barrier to deployment on resource-constrained edge devices, such as drones or surveillance cameras. This creates a tension between model size and representation power: lightweight models often struggle to distinguish foreground crowds from complex backgrounds without the capacity of their heavier counterparts.

To address these challenges simultaneously, this paper proposes a tiny Transformer-based Crowd counting framework, termed as TCFormer, a Vision Transformer with only 5 million parameters, which is an order of magnitude smaller than existing counters yet is trained end-to-end solely with image-level counts. Designing such an effective yet tiny model requires addressing the above challengers. The main contributions are fourfold:
%In this paper, we ask a challenging question: \textit{Can we achieve competitive counting performance with an ultra-lightweight model under the weakest supervision?}

%We answer this with \textbf{TCFormer} (Tiny Crowd Former), 
\begin{enumerate}
	\item This paper introduces an ultra-lightweight model architecture centered around a TinyViT backbone with only 5 million parameters. This design offers powerful feature extraction capabilities while maintaining extreme efficiency, making our model ideal for on-device inference.
	
	\item A Learnable Density Weighted Averaging (LDWA) module has been proposed to aggregate local features by weighting them according to their predicted density significance. This forces the network to adaptively focus on informative regions, enabling it to handle spatially varying crowd densities more effectively. The LDWA module enables the network to implicitly identify the location of the crowd in weakly-supervised settings without any location annotations. 
	
	\item This paper proposes a density-level classification loss function that categories image-level crowd count into crowd density levels while providing additional supervisory signals. These signals enhance the model's ability to perceive crowd density across diverse scenarios, thereby offering supplementary training signals to regulate weakly supervised learning processes.
	
	\item Using the above strategies, TCFormer, an ultra-lightweight, weakly-supervised transformer-based framework for crowd counting is constructed. Extensive experiments on four challenging benchmarks: ShanghaiTech A/B, UCF-QNRF, and NWPU demonstrate that our method achieves a superior trade-off between model complexity and counting accuracy. Specially, TCFormer not only outperforms prior weakly-supervised methods but also rivals completely supervised heavy networks, which further support the feasibility of our method in achieving robust performance without any dense annotations, while being capable of real-time inference on edge hardware with only 5 million parameters.
%	\item \textbf{Efficient Feature Extraction:} Standard ViTs quickly exceed parameter budgets when token dimensions grow. We adopt \textbf{TinyViT}~\cite{Wu2022} as our backbone. Unlike standard isotropic Transformers, TinyViT utilizes a hierarchical structure and distillation from larger teachers, enabling our model to capture robust semantic representations with minimal computational overhead.
%	
%	\item \textbf{Spatial Awareness without Labels:} In weakly-supervised settings, na\"ively pooling spatial tokens squashes informative high-density regions into negligible signals. To address this, we introduce a \textbf{Learnable Density-Weighted Averaging (LDWA)} module. This differentiable mechanism dynamically predicts a density attention map to re-weight local features before aggregation. Crucially, LDWA allows the network to implicitly learn "where" the crowd is without any location annotations.
%	
%	\item \textbf{Regularized Supervision:} To prevent the model from converging to trivial solutions (e.g., predicting random values that sum to the correct count), we propose an auxiliary \textbf{Density-Level Classification Loss}. We discretize the continuous count into ordinal density grades (e.g., light, moderate, heavy). This provides a coarser, yet stable, supervisory signal that enhances the model's ability to discriminate between scenes of varying congestion.
\end{enumerate}

The structure of this paper is organized as follows. Section~\ref{sec2} provides an overview of the current research on crowd counting. Section~\ref{sec3} explains the theory of the proposed methodologies, the experimental results and analysis of which are then detailed in Section~\ref{sec4}. Section~\ref{sec5} concludes the main work and contributions.

\section{Related Work}\label{sec2}
This section reviews related work in crowd counting, which broadly categorizes into three key aspects: fully supervised methods, weakly supervised approaches, and lightweight architectures.

\subsection{Fully-Supervised Crowd Counting}
Early crowd counting research employed detection-based approaches~\cite{li2008}, hand-crafted features~\cite{chen2012feature}, or Gaussian-process regression~\cite{chan2009}, which struggled significantly in high-density scenarios. Motivated by the success of Convolutional Neural Networks across computer vision, crowd counting rapidly moved away from hand-crafted pipelines in favor of CNN-based density-map regression. To address scale variation, researchers developed multi-column architectures~\cite{zhang2016single}, dilated convolutions~\cite{li2018csrnet}, and switching CNN ~\cite{sam2017}. 

The recent evolution has seen Vision Transformers (ViTs) demonstrating superior capability in capturing long-range dependencies, achieving state-of-the-art accuracy~\cite{hu2023, wang2024, jia2025}. More recent methods, such as APGCC~\cite{chen2024} and DLPTNet~\cite{chen2024privacy}, have further advanced the state-of-the-art through increasingly sophisticated attention mechanisms and transformer architectures. 

Of course, these approaches remain anchored in fully supervised paradigm, typically requiring training on large-scale datasets with dense, pixel-level annotations, such as ShanghaiTech~\cite{zhang2016single}. While they deliver high accuracy, they require pixel-level labels, which are labour-intensive and expensive to acquire. 

\subsection{Weakly-Supervised Crowd Counting}
To reduce annotation costs, Weakly Supervised Counting utilizes only global image-level counts, representing a promising alternative to fully supervised paradigms and addressing the fundamental challenge of annotation expense. Early approaches explored the sorting-based ranking~\cite{yang2020}, the crowd scale and distribution framework~\cite{fan2025}, and the dispersed attention mechanism~\cite{lei2025}. More recent methods~\cite{gao2025survey} involves leveraging large pre-trained foundation models for zero-shot or few-shot counting. For instance, CrowdCLIP~\cite{liang2023} adapts Vision Language Models to regress counts without location labels, and AdaSEEM~\cite{wan2024} uses the Segment Anything Model for crowd counting. Additional innovations include self-supervised methods~\cite{lin2025semi} and unsupervised methods for crowd counting~\cite{liu2020}.

While these foundation model-based approaches reduce the need for training labels, they introduce a new bottleneck in the form of large-scale model size. Deploying a CLIP or SAM backbone on edge devices is often infeasible. Furthermore, the self-supervised and unsupervised crowd counting methods still lag behind their supervised counterparts in terms of accuracy.

%In contrast, our work adheres to the weak supervision paradigm but utilizes a specialized lightweight backbone, avoiding the computational overhead of Large Vision Models (LVMs).
%
%While significantly reducing annotation dependency, these methods often trade off some accuracy compared to fully-supervised counterparts and face challenges in complex, high-density scenarios. Most existing weakly-supervised approaches employ heavy backbones to compensate for lack of spatial supervision, failing to address computational efficiency for edge deployment.

%Recently, transformer-based lightweight models have also emerged. \textbf{LEDCrowdNet}~\cite{yi2023effective} integrates MobileViT with a lightweight decoder, and \textbf{TinyCount}~\cite{lee2024tinycount} achieves extreme efficiency with only 0.06M parameters. However, a critical limitation persists across these lightweight works: they predominantly rely on \textit{full supervision} (dot-annotations). Our \textbf{TCFormer} distinguishes itself by being both ultra-lightweight (using TinyViT~\cite{Wu2022}) and weakly-supervised, effectively combining the efficiency of models like TinyCount with the data-efficiency of methods like CrowdCLIP.
\subsection{Lightweight Crowd Counting}
The demand for real-time edge analytics has prompted research into lightweight counting architectures, as comprehensively reviewed by Cheng et al.~\cite{cheng2025lw}. One stream of research focuses on compact CNN designs: GAPNet~\cite{guo2023ghost} employs GhostNet modules with lightweight attention, while Lw-Count~\cite{liu2022lw} optimizes encoding-decoding structures for efficiency. EdgeCount~\cite{shen2024lw} utilizes knowledge distillation to transfer density map knowledge to smaller student networks.

Recently, transformer-based lightweight models have also emerged. LEDCrowdNet~\cite{yi2023lw} integrates MobileViT with a lightweight decoder, and TinyCount~\cite{lee2024tc} achieves extreme efficiency with only 0.06M parameters. Although these lightweight methods achieve high FPS and low parameter counts, they typically struggle to maintain accuracy in dense, complex scenes compared to heavyweight models, and predominantly rely on full supervision.

In summary, existing literature reveals a persistent trilemma: high-performance models are either computationally intensive, rely on dense spatial supervision, or suffer substantial accuracy degradation when adopting lightweight architectures or weakly supervised paradigms, and the crowd counting paradigm lacks solutions that simultaneously address efficiency, accuracy, and annotation cost in a balanced framework. To the best of our knowledge, TCFormer represents the first framework that concurrently addresses three crucial aspects.

\begin{figure*}[]
	\centering
	\includegraphics[width=0.9\textwidth]{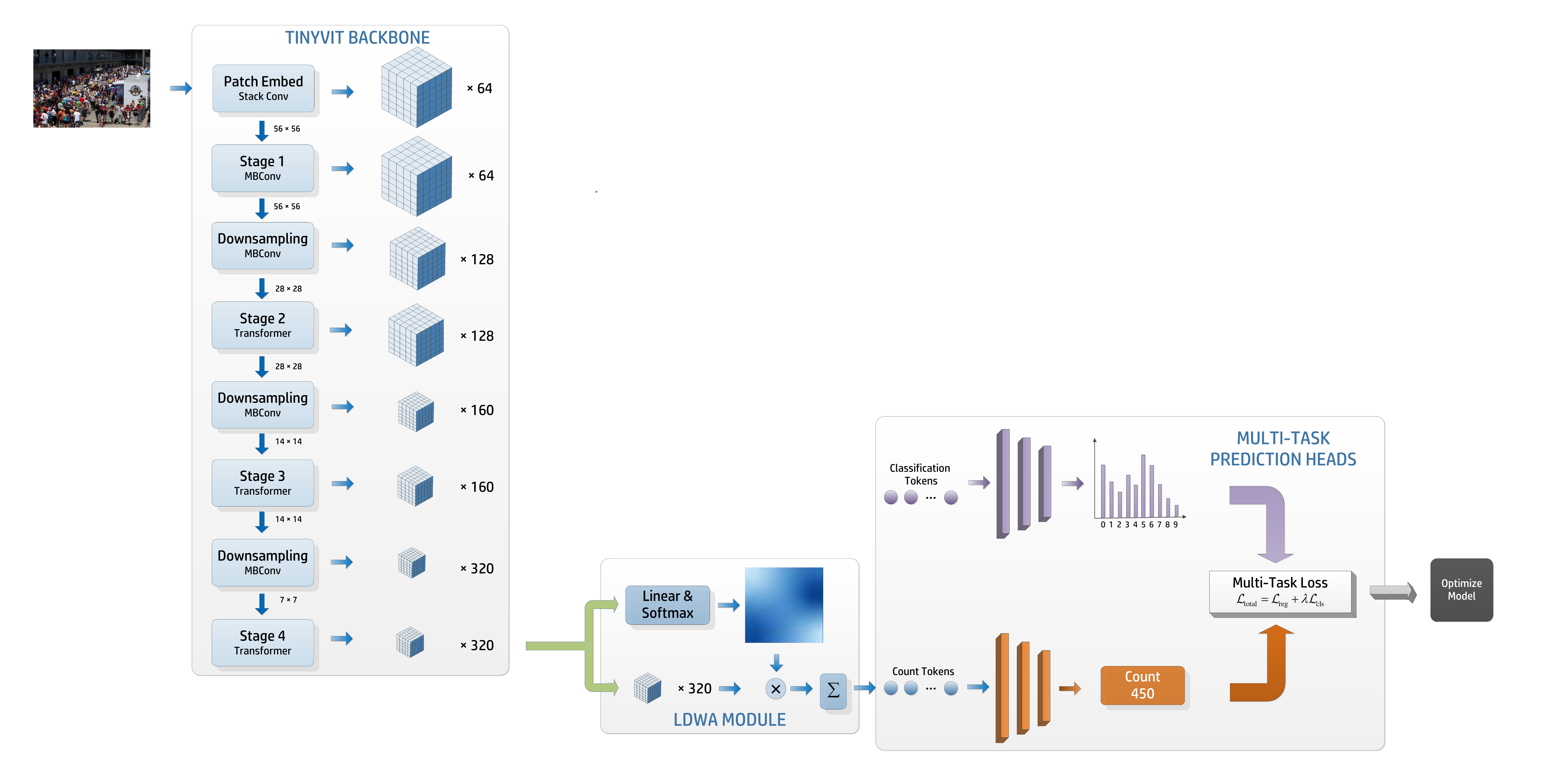}
	\caption{The crowd counting framework: TCFormer.}
	\label{fig:framework}
\end{figure*}

\section{Methodology}\label{sec3}
In this paper, an ultra-lightweight crowd counting framework: TCFormer is designed for crowd counting on edge deployment scenes. The TCFormer addresses the critical challenges of crowd counting through an ultra-lightweight Transformer architecture operating under weak supervision. As illustrated in Figure~\ref{fig:framework}, the model consists of three integral components: (1) A hierarchical TinyViT backbone for efficient, semantically meaningful feature extraction; (2) a Learnable Density-Weighted Averaging module that aggregates token features adaptively based on density-aware importance without requiring dense annotations; and (3) a multi-task prediction framework consisting of a count regression head and a density-level classification head driven by a learnable global prior. The auxiliary classification head provides complementary supervision to regulate the training process. The entire system is trained end-to-end using only image-level count annotations.

\subsection{TinyViT Backbone}
	
This paper adopts TinyViT~\cite{wu2022tvit} as our backbone due to its exceptional parameter efficiency and strong representation capabilities. Given an input image ${\emph{\textbf{I}}} \in {\emph{\textbf{R}}}^{H \times W \times 3}$, TinyViT produces a four-stage pyramid of token maps $\{{\emph{\textbf{T}}}_{s}\}_{s=1}^{4}$, where ${\emph{\textbf{T}}}_{s} \in {\emph{\textbf{R}}}^{\frac{H}{2^{s+1}} \times \frac{W}{2^{s+1}} \times C_{s}}$. This hierarchical structure captures both fine-grained texture details and high-level semantic context.

Given an input image ${\emph{\textbf{I}}} \in {\emph{\textbf{R}}}^{H \times W \times 3}$, the backbone produces semantic feature maps:

\begin{equation}
	%\emph{\textbf{F}} = \mathcal{T}(\emph{\textbf{I}}; \theta_{\text{tiny}}) \in \emph{R}^{h \times w \times d}
	\mathbf{\emph{T}}=\mathcal{F}(\emph{\textbf{I}},\theta_{\text{tiny}})\in\emph{\textbf{R}}^{h\times w\times d}
\end{equation}
where $\mathcal{F}$ denotes the TinyViT transformer with parameters $\theta_{\text{tiny}}$, and $(h,w,d)$ represent the reduced spatial dimensions and feature dimensionality respectively. 

TinyViT adopts a hierarchical design that explicitly integrates convolutional MBConv blocks and Transformer-based self-attention across multiple stages. In the early stages, lightweight MBConv layers efficiently capture fine-grained local textures and spatial structures, which are critical for representing densely regions. As the network depth increases, Transformer blocks are progressively introduced to model long-range dependencies and aggregate global contextual information through self-attention. 

Specifically, TinyViT produces a four-stage pyramid of token maps with gradually decreasing spatial resolution and increasing semantic abstraction. To maximize computational efficiency while retaining strong representational capacity, This paper utilizes the feature map from the final stage as the global crowd representation. Compared to multi-scale feature fusion strategies that introduce additional computational overhead, leveraging the final-stage features provides a compact yet semantically rich representation that is well suited for crowd counting and density-level prediction.

Let the output of the fourth stage be denoted as ${\emph{\textbf{T}}}_4 \in {\emph{\textbf{R}}}^{\frac{H}{32} \times \frac{W}{32} \times C_4}$, where $C_4$ is the channel depth of the final stage. Then, it will be as the encoder tokens for crowd counting in next processing procedures.

%We apply a $1\times1$ convolution to project these features to a unified dimension $d=256$, resulting in our feature map ${\emph{\textbf{F}}}$:
%\begin{equation}
%	{\emph{\textbf{F}}} = \text{Conv}_{1\times1}({\emph{\textbf{T}}}_4) \in {\emph{\textbf{R}}}^{h \times w \times d}
%\end{equation}
\subsection{Learnable Density-Weighted Averaging Module}
Image-level supervision provides only a global headcount, lacking any spatial clues about the actual standing positions of pedestrians. Consequently, the network receives no hints about the highly non-uniform distribution of crowd, yet this distribution is the key factor determining counting accuracy due to densely crowd areas contribute far more to the total count than sparsely populated regions. To internalize this prior knowledge, a Learnable Density-Weighted Averaging (LDWA) module is proposed to re-weight every token proportional to its estimated crowding level. LDWA automatically promotes the influence of congested areas while suppressing sparse ones, allowing the ultra-light model to focus on the most critical locations for final counting.

%A critical limitation of weakly-supervised learning is the lack of spatial labels. Standard global average pooling treats background (e.g., sky) and foreground (crowd) equally, diluting the counting signal. We propose the \textit{Density-Guided Aggregation} (DGA) module, 
Let $T_4$ denote the feature tokens extracted from the backbone. The proposed mechanism operates through the following mathematical formulation:

\begin{equation}
	\emph{\textbf{S}} = \emph{\textbf{T}}_{4} \emph{\textbf{W}}_d^\top + \emph{\textbf{b}}_d
\end{equation}
where $\emph{\textbf{W}}_d \in \emph{\textbf{R}}^{1 \times d}$ and $\emph{\textbf{b}}_d \in \mathbf{R}$ are learnable parameters projecting each token to a scalar density score $\emph{\textbf{S}} \in \emph{\textbf{R}}^{N \times 1}$. $N = h \times w $

Next, these scores are converted into a normalized weight distribution suitable for aggregation, this paper uses the Softmax function across the spatial dimensions ($h \times w$):
\begin{equation}
	%\alpha_{ij} = \frac{\exp(s_{ij})}{\sum_{x=1}^{h}\sum_{y=1}^{w} \exp(s_{xy})}
	\emph{\textbf{W}} = \text{softmax}(\emph{\textbf{S}}) = \frac{\exp(\emph{\textbf{S}}_i)}{\sum_{j=1}^{N} \exp(\emph{\textbf{S}}_j)}
\end{equation}
This operation ensures that $\sum_{j} \omega_{j} = 1$, allowing the model to assign relative importance to different regions.

The globally aggregated feature representation $\emph{\textbf{T}}_{count}$ is obtained by computing the weighted sum of the input features:
\begin{equation}
	%{\emph{\textbf{g}}} = \sum_{i=1}^{h}\sum_{j=1}^{w} \alpha_{ij} \cdot {\emph{\textbf{F}}}_{ij}
	\emph{\textbf{T}}_{count} = \sum_{j=1}^{N} \emph{\textbf{W}}_j \emph{\textbf{T}}_4^j
\end{equation}

The complete differentiable transformation can be compactly expressed as:
\begin{equation}
	\emph{\textbf{T}}_{count} = \sum_{j=1}^{N} \left[ \text{softmax}\left( \emph{\textbf{T}}_{4} \emph{\textbf{W}}_d^\top + \emph{\textbf{b}}_d \right)_j \emph{\textbf{T}}_4^j \right]
\end{equation}

By optimizing this process end-to-end, the network implicitly learns to assign higher weights $\emph{\textbf{w}}_j$ to high-density regions, effectively filtering out background noise and focusing the count regression on the relevant crowd features, thus a learnable attention mechanism that dynamically re-weights local features is build.

%First, a scalar density importance score $s_{ij}$ for each spatial location is computed:
%\begin{equation}
%	s_{ij} = \sigma({\emph{\textbf{W}}}_d \cdot \phi({\emph{\textbf{F}}}_{ij}) + b_d) \in (0, 1)
%\end{equation}
%where $\phi(\cdot)$ denotes a non-linear projection ($\text{BN} \to \text{ReLU}$), ${\emph{\textbf{W}}}_d$ is a learnable weight vector, and $\sigma$ is the sigmoid function.
%
%Next, we normalize these scores to obtain a spatial attention map $\alpha$:
%\begin{equation}
%	\alpha_{ij} = \frac{s_{ij}}{\sum_{x=1}^{h}\sum_{y=1}^{w} s_{xy} + \epsilon}
%\end{equation}
%where $\epsilon=10^{-6}$ ensures numerical stability. The final global feature vector ${\emph{\textbf{g}}}$ is computed as the density-weighted sum:
%\begin{equation}
%	{\emph{\textbf{g}}} = \sum_{i=1}^{h}\sum_{j=1}^{w} \alpha_{ij} \cdot {\emph{\textbf{F}}}_{ij}
%\end{equation}
%This mechanism forces the network to implicitly localize crowd regions by assigning higher weights $\alpha_{ij}$ to areas with high density, effectively filtering out background noise.

\subsection{Multi-Task Prediction Heads}

In order to exploit various supervision signals to compensate for the absence of dense annotations, this paper designs a multi-task learning framework in which count regression and density-level classification provide complementary regulatory constraints, effectively guiding model training toward more accurate crowd counting predictions.

\subsubsection{Count Regression Head}
The count regression head transforms the count-aware global feature representation into
a precise crowd count estimation. Given the aggregated feature
$\emph{\textbf{T}}_{count} \in \emph{\textbf{R}}^{1 \times d}$, and the regression head performs a linear projection:

\begin{equation}
	y_{\text{reg}} = \emph{\textbf{W}}_{reg} \cdot \emph{\textbf{T}}_{count} + b_{\text{reg}}
\end{equation}
where $\mathbf{W}_{\text{reg}} \in \emph{\textbf{R}}^{1 \times d}$ and
$b_{\text{reg}} \in \emph{R}$ are learnable parameters, and
$y_{\text{reg}} \in \emph{R}$ denotes the predicted crowd count.

This regression branch is optimized using a count regression loss and focuses on capturing
fine-grained, image-dependent crowd variations.

\subsubsection{Density-Level Classification Head}
In addition to count regression, this paper introduces a density-level classification head to
enhance the model's robustness across diverse crowd scales.
Instead of relying on image features, this head is driven by a learnable
tokens, which serves as a global trainable vector that are defined as $\emph{\textbf{T}}_{density} \in \emph{\textbf{R}}^{1 \times d}$, which is directly fed into the classification head. Notably, the token does not participate in Transformer layers nor interact with image tokens via attention. It is optimized solely through back-propagation from the
density-level classification loss.

In this paper, the continuous crowd count range is discretized into $K$ density levels using uniform quantization:
\begin{equation}
	\hat{y}_{\text{cls}} = \left\lfloor \frac{K \cdot \min(c, c_{\text{max}})}{c_{\text{max}}} \right\rfloor,
\end{equation}
where $c$ is the ground-truth count, $c_{\text{max}}$ is the maximum count in the dataset,
and $\hat{y}_{\text{cls}}\in\{0, 1, \ldots, K-1\}$ denotes the corresponding density level.

The classification head applies a linear projection followed by softmax normalization:
\begin{equation}
	y_{\text{cls}} = \text{softmax}(\emph{\textbf{W}}_{cls} \cdot \emph{\textbf{T}}_{density} + b_{\text{cls}})
\end{equation}
where $\emph{\textbf{W}}_{cls} \in \emph{\textbf{R}}^{K \times d}$ and
$\emph{\textbf{b}}_{cls} \in \emph{\textbf{R}}^{K}$ are learnable parameters, and
$y_{\text{cls}} \in \emph{\textbf{R}}^{1 \times K}$ represents the predicted density levels.

Although the classification token is not conditioned on individual images, it functions as a
learnable global prior that captures dataset-level statistical regularities,
such as crowd density ranges and scale distributions. By jointly optimizing
the regression and classification objectives, the classification branch
implicitly regularizes the counting model and improves prediction stability and
generalization.

%\subsubsection{Count Regression Head}
%The count regression head transforms the global feature representation into a precise crowd count estimation. Given the aggregated feature ${\emph{\textbf{g}}}$, the regression head performs a linear projection:
%
%\begin{equation}
%	y_{reg} = \mathbf{W}_{reg} \cdot {\emph{\textbf{g}}} + b_{reg}
%\end{equation}
%where $\mathbf{W}_{reg} \in \mathbf{R}^{1 \times d}$ and $b_r \in \mathbf{R}$ are learnable parameters, and $\hat{c} \in \mathbf{R}^{B \times 1}$ represents the predicted crowd counts.
%
%\subsubsection{Density-Level Classification Head}
%To enhance the model's discrimination across varying crowd levels, an density-level classification head. The continuous count range is discretized into $K$ density levels using a uniform quantization scheme:
%\begin{equation}
%	y_{\text{density}} = \left\lfloor \frac{K \cdot \min(c, c_{\text{max}})}{c_{\text{max}}} \right\rfloor
%\end{equation}
%
%where $c$ is the ground-truth count, $c_{\text{max}}$ is the dataset maximum count, and $y_{\text{density}} \in \{0, 1, \ldots, K-1\}$ represents the discretized density level.
%
%The classification head processes the global feature through a linear layer followed by softmax normalization:
%
%\begin{equation}
%	\hat{\mathbf{p}} = \text{softmax}(\mathbf{W}_c \cdot \mathbf{f}_{\text{global}} + \mathbf{b}_c)
%\end{equation}
%
%where $\mathbf{W}_c \in \mathbf{R}^{K \times d}$ and $\mathbf{b}_c \in \mathbf{R}^{K}$ are learnable parameters, and $\hat{\mathbf{p}} \in \mathbf{R}^{B \times K}$ represents the predicted probability distribution over density levels.
\begin{table*}[]
	\centering
	\caption{Summary of the datasets used in this study. The ShanghaiTech (Parts A/B) dataset are referred to as Part A and Part B.}\label{tab:dataset}
	\renewcommand{\arraystretch}{1.2}
	\setlength{\tabcolsep}{8pt}
	\begin{tabular}{lccccccccc}
		\toprule
		\textbf{Dataset} & \multicolumn{5}{c}{\textbf{Image Statistics}} &
		\multicolumn{3}{c}{\textbf{Crowd Counting Statistics}} \\
		\cmidrule(lr){2-6} \cmidrule(lr){7-9}
		& \textbf{Images} & \textbf{Train} & \textbf{Val} & \textbf{Test} & \textbf{Resolution} &
		\textbf{Min} & \textbf{Max} & \textbf{Mean} \\
		\midrule
		Part A       & 482  & 300  & -- & 182  & $589{\times}868$   & 33   & 3139   & 501 \\
		Part B      & 716  & 400  & -- & 316  & $768{\times}1024$  & 9    & 578    & 123 \\
		%UCF\_CC\_50~\cite{62}  & 50   & 40   & -- & 10   & --                & 94   & 4543   & 1280 \\
		UCF-QNRF     & 1535 & 1201 & -- & 334  & $2013{\times}2902$ & 49   & 12865  & 815 \\
		NWPU-Crowd   & 5109 & 3190 & 500 & 1500 & $2191{\times}3209$ & 0    & 20033  & 418 \\
		%WorldExpo'10~\cite{65} & 3920 & 3380 & -- & 600  & $576{\times}720$   & --   & --     & -- \\
		%JHU-CROWD++~\cite{66}  & 4372 & 2272 & 500 & 1600 & $910{\times}1430$  & 0    & 25791  & 346 \\
		\bottomrule
	\end{tabular}
\end{table*}
\subsection{Overall Loss}
The overall training loss combines both regression and classification losses through a multi-task learning framework, which is introduced below:

\noindent \textbf{Regression Loss:}  the \textbf{Smooth L1 Loss} is employed for robust count estimation:

\begin{equation}
	\mathcal{L}_{\text{reg}} = \frac{1}{M} \sum_{i=1}^{M} 
	\begin{cases} 
		0.5(\hat{y}_{reg}^i - y_{reg}^i)^2, & \text{if } |\hat{y}_{reg}^i - y_{reg}^i| < 1 \\
		|\hat{y}_{reg}^i - y_{reg}^i| - 0.5, & \text{otherwise}
	\end{cases}
\end{equation}
$\hat{y}_i$ and $y_i$ denote the ground-truth and predicted crowd counts of the $i$-th image, respectively. $M$ represents the amount of images in the training dataset.

\noindent \textbf{Classification Loss:} The density-level classification employs binary
cross-entropy loss:
%	\mathcal{L}_{\text{cls}} = -\frac{1}{B} \sum_{i=1}^{B} \sum_{k=0}^{K-1} \mathbf{1}\{y_{\text{density}}^{(i)} = k\} \cdot \log(\hat{p}_{i,k})
\begin{equation}
	\mathcal{L}_{\text{cls}} = -\sum_{i=1}^{M} \left[ \hat{y}^{i}_{\text{cls}} \log(y_{cls}^i) + (1 - \hat{y}^{i}_{\text{cls}}) \log(1 - y_{cls}^i)) \right]
\end{equation}
where $M$ denotes the number of images, $\hat{y}^{i}_{\text{cls}}$ is the ground-truth classification label of the $i$-th sample, and $y_{\text{cls}}^i$ represents the predicted probability produced by the classification head.

%To regularize the training and prevent trivial solutions (e.g., predicting the mean count), we introduce an auxiliary density classification task. We discretize the crowd count range into $K=3$ ordinal grades: \textit{Sparse} $[0, 50)$, \textit{Moderate} $[50, 500)$, and \textit{Congested} $[500, \infty)$.
Finally, the overall training loss is computed as:
\begin{equation}
	\mathcal{L}_{\text{total}} = \mathcal{L}_{\text{reg}} + \lambda \mathcal{L}_{\text{cls}}
\end{equation}
where hyperparameter $\lambda=0.001$ balances the contributions of both loss terms, enabling joint optimization of count accuracy and density-aware feature learning under weak supervision constraints.
%
%A classification head predicts the logits ${\emph{\textbf{p}}} \in {\emph{\textbf{R}}}^{K}$. We use the \textbf{Focal Loss} to handle class imbalance, as extreme density images are often rare:
%\begin{equation}
%	\mathcal{L}_{\text{cls}} = -\sum_{k=1}^{K} (1 - p_k)^\gamma \log(p_k)
%\end{equation}
%where $p_k$ is the predicted probability for the correct class and $\gamma=2$ is the focusing parameter.

\section{Experiments and Results}\label{sec4}
This section provides details of the experimental setup and results.

\subsection{Experimental Setup}
\label{subsec:setup}

\subsubsection{Datasets}
This paper evaluates the proposed method on four mainstream benchmarks covering diverse scene types and crowd densities:
\begin{itemize}
	%	\item \textbf{ShanghaiTech Part A (SHA)}~\cite{zhang2016single}: 482 images with 244,167 annotations (300 training, 182 testing), averaging 501 heads per image.
	%	\item \textbf{ShanghaiTech Part B (SHB)}~\cite{zhang2016single}: 716 images with 88,498 annotations (400 training, 316 testing), averaging 123 heads per image.
	%	\item \textbf{UCF-QNRF (QNRF)}~\cite{idrees2018composition}: 1,535 high-resolution images with 1.25 million annotations (1,201 training, 334 testing), averaging 815 heads per image.
	%	\item \textbf{JHU-Crowd++ (JHU++)}~\cite{sindagi2022jhu}: 4,372 images with 1.51 million annotations, notable for diverse weather conditions and severe occlusion (2,272 training, 500 validation, 1,600 testing).
	%	\item \textbf{NWPU-Crowd (NWPU)}~\cite{wang2021nwpu}: 3,609 images (total 5,109 including 1,500 negative samples) with extreme density variations (0--20,033 persons per image).
	\item ShanghaiTech Dataset \cite{zhang2016single}
	This dataset contains two subsets: Part~A and Part~B. 
	Part~A includes 482 images, with 300 allocated for training and 182 for testing. 
	This subset includes highly congested scenes, where the number of people per image ranges from 33 to 3193. 
	Part~B consists of 716 images, including 400 training images and 316 testing images. 
	Compared with Part~A, Part~B is significantly sparser, with crowd counts varying between 9 and 578.
	
	\item  UCF-QNRF Dataset \cite{idrees2018}
	This dataset contains 1,535 high-density crowd images, of which 1201 are used for training and 334 for testing. It represents one of the densest datasets in crowd counting, with the number of people per image ranging from 49 to 12,865.
	
	\item NWPU-Crowd Dataset \cite{wang2020nwpu}
	This dataset comprises 5,109 images, which is a large-scale benchmark for both crowd counting and localization, including diverse scenes with significant variations in density and illumination. The number of people per image ranges from 0 to 20,003, making it highly challenging for model training.
\end{itemize}

The detailed information of these datasets are listed in the Table~\ref{tab:dataset}.
%, and some samples on these datasets are shown in Figure~\ref{tab:dataset}

\subsubsection{Hardware Environment}
The experimental platform specifications are detailed in Table~\ref{tab:hardware_env}, providing detailed information about hardware and software configurations used in our experiments. This high-performance computing setup ensures efficient training and inference for the proposed crowd counting models.

\begin{table}[!h]
	\centering
	\caption{Hardware and Software Environment Specifications}
	\label{tab:hardware_env}
	\begin{tabular}{@{}ll@{}}
		\toprule
		\textbf{Component} & \textbf{Specification} \\
		\midrule
		\textbf{CPU} & 2 $\times$ Kunpeng 920 \\
		\textbf{GPU} & 4 $\times$ NVIDIA A100 \\
		\textbf{Memory} & 256GB DDR4 RAM \\
		\textbf{Operating System} & Ubuntu 22.04 LTS \\
		\textbf{Deep Learning Framework} & PyTorch 2.2.2 \\
		\bottomrule
	\end{tabular}
\end{table}

%, with the A100 GPUs providing accelerated computation through tensor cores and high memory bandwidth.

\subsubsection{Implementation Details}
The detailed training configurations for the TCFormer are summarized in Table~\ref{tab:tcformer_config}. These settings have been empirically validated to guarantee stable convergence and discriminative feature learning for crowd-counting tasks.

\begin{table}[h]
	\centering
	\caption{TCFormer Training Protocol}
	\label{tab:tcformer_config}
	\small
	\begin{tabular}{@{}ll@{}}
		\toprule
		\textbf{Hyper-parameter} & \textbf{Value} \\
		\midrule
		Optimizer & Adam ($\beta_{1}=0.9, \beta_{2}=0.999$) \\
		Peak Learning Rate & $5 \times 10^{-5}$ \\
		Learning Rate Scheduler & MultiStepLR \\
		Weight Decay & $5 \times 10^{-4}$ \\
%		Momentum (EMA) & 0.95 \\
		Mini-batch Size & 16 \\
		Training Epochs & 500 \\
		Input Resolution & $224 \times 224$ \\
		Data Augmentation & Random horizontal flip ($p=0.5$) \\
		\bottomrule
	\end{tabular}
\end{table}

%optimized using the Adam optimizer with a learning rate of $1\times10^{-5}$ for backbone parameters and $1\times10^{-4}$ for others. The batch size is set to 8. Data augmentation includes random scaling (factors [0.7, 1.3]), random cropping ($256\times256$ for QNRF, JHU++, NWPU; $128\times128$ for SHA, SHB), and horizontal flipping (0.5 probability).

%For the Scale-based Active Learning (SAL) strategy, we utilize scaling factors of 0.8 and 1.2 to generate multi-scale images. In the 40\% annotation budget setting, initialization is 20\% random labels, adding 5\% per active learning cycle.

\subsubsection{Evaluation Metrics}
The key metrics, including total parameters, Mean Absolute Error (MAE) and Root Mean Squared Error, are adopted to fully assess the comprehensive prediction performance of all models concerning computational complexity and detection accuracy. This evaluation reflects the strengths and weaknesses of these models in crowd counting problem by comparing the experimental results across these metrics. Additionally, it can comprehensively evaluate the effectiveness of the proposed methods in enhancing the crowd counting capability.

\paragraph{Accuracy Metrics}
The widely used quantitative metrics in crowd counting: \emph{Mean Absolute Error} (MAE) and \emph{Root Mean Squared Error} (RMSE), the computing equation of which is shown below. 

Given a test set with $M$ images, predicted counts $y_{reg}^i$, and ground-truth counts $\hat{y}_{reg}^i$, the metrics are defined as
\begin{equation}
	\mathrm{MAE} = \frac{1}{M} \sum_{i=1}^{M} \left| \hat{y}_{reg}^i - y_{reg}^i \right|,
\end{equation}
\begin{equation}
	\mathrm{RMSE} = \sqrt{ \frac{1}{M} \sum_{i=1}^{M} \left( \hat{y}_{reg}^i - y_{reg}^i \right)^{2} }.
\end{equation}

MAE reflects the average absolute deviation between predictions and ground truth, while RMSE mphasizes larger errors through squared penalization, making it more sensitive to outliers and severe prediction errors.

\begin{table*}[ht]
	\centering
	\caption{Comparison of recent state-of-the-art crowd counting methods on benchmark datasets. Methods are grouped by supervision type: Fully Supervised (FS) and Non-/Weakly-Supervised (NF). ``Params (M)'' reports the total number of trainable parameters. ShanghaiTech Part A/B are listed as Part A and Part B. Values are taken from the referenced literature (see main text) and the supplied extract.}
	\label{tab:comparison_final_revised}
	\scriptsize
	\begin{tabular}{l c c c c c c c c c c c}
		\toprule
		\multirow{2}{*}{\textbf{Method}} &
		\multirow{2}{*}{\textbf{Year}} &
		\multirow{2}{*}{\textbf{Type}} &
		\multirow{2}{*}{\textbf{Params (M)}} &
		\multicolumn{2}{c}{\textbf{Part A}} &
		\multicolumn{2}{c}{\textbf{Part B}} &
		\multicolumn{2}{c}{\textbf{UCF-QNRF}} &
		\multicolumn{2}{c}{\textbf{NWPU}} \\
		\cmidrule(lr){5-6}  \cmidrule(lr){7-8}  \cmidrule(lr){9-10} \cmidrule(lr){11-12}
		& & & & MAE & RMSE & MAE & RMSE & MAE & RMSE & MAE & RMSE \\
		\midrule
		\multicolumn{12}{l}{\textbf{Heavyweight Models:}} \\
		SFCN           & 2021 & FS  & 38.6 & 64.8 & 107.5 & 7.6 & 13.0  & 102.0 & 171.4  & 105.7   & 424.1     \\
		%STNet \cite{116}              & 2022 & FS  & 15.56 & 52.9 & 83.6  & 6.3 & 10.3  & 87.9 & 166.4  & --    & --     \\
		SRRNet              & 2023 & FS  & 66.14 & 60.8 & 103.0 & 7.4  & 13.6  & 89.5  & 162.9  & --  & --  \\
		%PET \cite{117}                & 2023 & FS  & 56.14 & 60.0 & 98.6  & 10.1 & 17.4  & 153.3 & 236.5  & --    & --     \\
		RAQNet             & 2024 & FS  & 42.77 & 59.0 & 101.2  & 9.0  & 15.4  & 106.5 & 186.1  & --    & --     \\
		DLPTNet             & 2024 & FS  & 110.90 & 58.4 & 95.0  & 9.3  & 15.6  & 121.0 & 225.8  & 103.3 & 421.9  \\
		SDANet              & 2024 & FS  & 56.50 & 54.9 & 90.4  & 7.1  & 12.0  & 107.3 & 195.5  & --    & --     \\
		\midrule
		\multicolumn{12}{l}{\textbf{Lightweight Models:}} \\
		TinyCount            & 2024 & FS  & 0.06  & 78.2 & 120.8 & 10.8  & 18.4  & 134.7  & 223.3  & --    & --     \\
		LMSNet               & 2024 & FS  & 0.73  & 62.9 & 108.4  & 8.2 & 13.5  & 110.7 & 178.7  & --    & --     \\
		LMSFFNet             & 2024 & FS  & 2.88  & 85.9 & 139.9 & 9.2  & 15.1  &  112.8 & 372.8  & --    & --     \\
		PDDNet               & 2024 & FS  & 1.10  & 72.6 & 112.2 & 10.3 & 17.0  & 130.2 & 246.6  & 91.5  & 381.0  \\
		RepMobileNet         & 2024 & FS  & 3.41  & 84.2 & 127.5 & 8.6  & 13.7  & 122.5  & 201.6  & --    & --     \\
		DHMoE         & 2025 & FS  & 5.86  & 59.2 & 96.1 & 11.0  & 19.6  & 132.1  & 253.6  & 118.0    & 481.1     \\
		\midrule
	    \multicolumn{12}{l}{\textbf{Non-Fully Supervised Models:}} \\
	    DACount               & 2022 & NF  & ---   & 82.5 & 123.2 & 10.9  & 19.1  & 115.1  & 193.5  & --    & --     \\
	    OT-M                  & 2023 & NF  & ---   & 80.1 & 118.5 & 10.8  & 18.2  & 113.1 & 186.7  & --    & --     \\
	    MRL                   & 2023 & NF  & ---   & 80.2 & 125.6 & 12.1 & 19.7  & 132.5 & 221.2  & 132.9  & 511.3  \\
	    Zhang et al.                & 2024 & NF  & ---   & 69.7 & 114.5 & 9.7  & 17.5 & 106.7  & 171.3  & 107.1  & 443.6  \\
		\midrule
		\textbf{TCFormer (Ours)}       & \textbf{2025} & \textbf{NF} & \textbf{5.52} & \textbf{64.541} & \textbf{110.052} & \textbf{8.155} & \textbf{14.024} & \textbf{93.566} & \textbf{162.694} & \textbf{79.467} & \textbf{360.580} \\
		\bottomrule
	\end{tabular}
\end{table*}

\paragraph{Efficiency Metrics}
To evaluate computational efficiency, this paper reports the number of model parameters, the required floating-point operations, inference latency throughput measured in frames per second.  

%
%We adopt both accuracy and efficiency metrics for comprehensive evaluation:
%\begin{itemize}
%	\item \textbf{Accuracy Metrics:}
%	\begin{itemize}
%		\item \textbf{Mean Absolute Error (MAE)}: $\frac{1}{N} \sum_{i=1}^{N} |C_i - GT_i|$
%		\item \textbf{Root Mean Squared Error (RMSE)}: $\sqrt{\frac{1}{N} \sum_{i=1}^{N} (C_i - GT_i)^2}$
%	\end{itemize}
%	\item \textbf{Efficiency Metrics:} Model parameters (Params, M), Floating Point Operations (FLOPs, G), Inference Latency (ms), and Frames Per Second (FPS) measured on $576\times768$ input with batch size 1.
%\end{itemize}

\subsubsection{Baseline Models}
To rigorously evaluate the effectiveness of the TCFormer for crowd counting, this paper conducts comparative experiments with recent crowd counters. According to their model complexity and supervision type, the baselines can be broadly divided into three main categories:

\begin{itemize}
\item \textbf{Fully Supervised (FS) Heavyweight Models:} 
SFCN~\cite{wang2021sfcn}, 
SRRNet~\cite{guo2023ITS}, 
RAQNet~\cite{zhai2024raq}, 
DLPTNet~\cite{chen2024privacy}, 
SDANet~\cite{wang2024sda}.

\item \textbf{Fully Supervised (FS) Lightweight Models:} 
TinyCount~\cite{lee2024tc}, 
LMSNet~\cite{xi2024}, 
LMSFFNet~\cite{yi2023lmsff}, 
PDDNet~\cite{liang2023pdd}, 
RepMobileNet~\cite{lin2024rep}, 
DHMoE~\cite{cheng2025dhmoe}.

\item \textbf{None-Fully Supervised (NF) Models:} 
DACount~\cite{lin2022Semi}, 
OT\text{-}M~\cite{lin2023otm}, 
MRL~\cite{wei2023mrl}, 
Zhang~et~al.~\cite{zhang2024boost}.
\end{itemize}

\subsection{Benchmark experiments}
\subsubsection{ShanghaiTech Part A \& Part B}
On the ShanghaiTech dataset, our method achieves competitive prediction performance in Parts A and B. 
For Part A, our approach outperforms several lightweight methods, achieving maximum improvements of 21.36\% (MAE) and 29.38\% (RMSE), highlighting its strong capability in handling dense crowd scenarios while maintaining a compact model size. 
However, when compared with other lightweight methods such as DHMoE, our method exhibits slightly inferior performance, with gaps of 5.341\% (MAE) and 13.952\% (RMSE), indicating a modest trade-off between computational efficiency and prediction accuracy. 
Furthermore, as a non-fully supervised model, our approach significantly outperforms unsupervised methods, demonstrating that a carefully designed architecture can markedly improve the accuracy of crowd density estimation with limited supervision. 
This validates the effectiveness of the proposed method.

For Part B, which contains sparser crowd scenes, our method achieves performance comparable to that of heavyweight models, with maximum margins of 0.755\% (MAE) and 2.024\% (RMSE). 
This demonstrates the adaptability of our approach to varying crowd densities and its ability to maintain strong prediction accuracy in low-density scenarios. 
Overall, these results suggest that our method strikes a balance between estimation accuracy and computational efficiency, making it suitable for both dense and sparse crowd counting tasks.

\subsubsection{QNRF}
On the QNRF dataset, which consists of ultra--high-resolution images and exhibits substantial variations in crowd density, the proposed method demonstrates strong generalization capability and achieves lower MAE and RMSE than most heavyweight models, indicating superior accuracy in large-scale and highly crowded scenes. Notably, only one MAE result is slightly higher than that of the most powerful heavyweight model, with the performance difference remaining within approximately 4.066\%, while all other MAE and RMSE metrics are superior. 
Compared with unsupervised approaches, the TCFormer significantly surpasses them, reducing MAE and RMSE by up to 38.934\% and 58.506\%, respectively.
Overall, these results demonstrate that the proposed model achieves a favorable balance between accuracy and generalization, learning robust and transferable representations that enable reliable crowd estimation under challenging, highly congested real-world scenarios.

\subsubsection{NWPU}
On the NWPU benchmark, the proposed approach achieves the lowest MAE and RMSE among all compared methods, outperforming both heavyweight and lightweight baselines by clear margins. 
Specifically, the TCFormer reduces MAE and RMSE by up to 20.42\% and 20.42\%, respectively, compared with the best-performing competing models. 
These consistent reductions in error metrics demonstrate improved precision in crowd estimation across scenes with large scale variation and complex backgrounds. These results highlight the strong generalization potential of our proposed methods, confirming its reliability and robustness for high-density, large-scale crowd estimation.

The counting performance is evaluated using MAE and RMSE. To facilitate a fair cross-dataset comparison and account for the varying crowd densities across benchmarks, this paper employs normalized MAE (nMAE) and normalized RMSE (nRMSE). These metrics are defined as follows:
\begin{equation}
	\mathrm{nMAE} = \frac{1}{M}\sum_{i=1}^{M} \frac{\mathrm{MAE}_i}{\overline{C}_i}, \\
	\mathrm{nRMSE} = \frac{1}{M}\sum_{i=1}^{M} \frac{\mathrm{RMSE}_i}{\overline{C}_i},
\end{equation}
where $\mathrm{MAE_i}$ and $\mathrm{RMSE_i}$ are the $\mathrm{MAE, RMSE}$ on the $i$-th dataset. $\overline{Y}_i$ denotes the mean ground-truth count of the $i$-th dataset. $M$ is the total number of test images.

By normalizing the absolute errors by the dataset's mean density, the resulting nMAE and nRMSE provide a relative measure of precision. This ensures that the performance gains observed on high-density datasets (UCF-QNRF) are comparable to those on lower-density benchmarks (ShanghaiTech Part B), providing a more holistic view of the model's scale-invariant robustness.
 
\begin{figure}[h]
	\centering
	\includegraphics[width=0.4\textwidth]{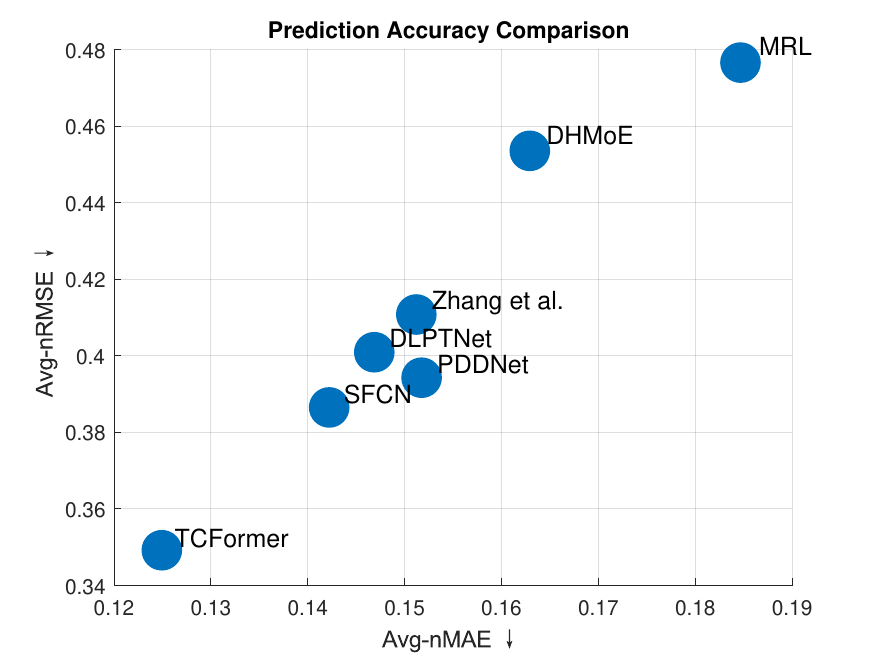}
	\caption{The nMAE \& nRMSE value of some models on benchmark datasets.}
	\label{FIG:pac}
\end{figure}

As shown in Fig.~\ref{FIG:pac}, the nMAE and nRMSE metrics of representative crowd counting models are presented. By evaluating performance across all benchmarks, including ShanghaiTech Part A and B, UCF-QNRF, and NWPU, the figure highlights the consistent superiority of our method.

Specifically, the TCFormer achieves the lowest values for both metrics across all datasets. Despite significant variations in average crowd density across these benchmark datasets, the consistent performance demonstrates the robustness of our architecture. While many existing lightweight models struggle with the extreme crowding scenarios in UCF-QNRF and NWPU datasets, our model maintains high accuracy. This success contributes to the hierarchical feature extraction of the TinyViT backbone and the adaptive focusing of the LDWA module, ensuring spatial features receive effective weighting across varying crowd scales. Furthermore, results in Figure~\ref{FIG:pac} demonstrate that our model remains competitive even against heavyweight baselines, achieving an good balance between counting accuracy and computational efficiency.

%The proposed TCFormer achieves the lowest values on both measures, indicating superior prediction performance compared to existing approaches. Notably, these results are attained with modest computational cost, underscoring the efficiency of the proposed framework.
\label{subsec:comparison}
%Table~\ref{{tab:results_accuracy}} presents counting accuracy on ShanghaiTech Part A/B, UCF-QNRF, JUH, and NWPU.  TCFormer significantly outperforms the lightweight baselines: e.g., it achieves a Part A MAE of 53.0 (RMSE 82.5), compared to 78.2 MAE (120.8 RMSE) for TinyCountand 67.1 MAE (110.4 RMSE) for GAPNet.  TCFormer’s performance approaches that of heavyweight models.  For instance, APGCC (18.7M params) reports 48.8 MAE on Part A, and our model’s 53.0 MAE is only slightly higher despite having $4\times$ fewer parameters.  On Part B, TCFormer attains 6.2 MAE (9.5 RMSE), close to APGCC’s 5.6 MAE.  Similar trends hold on UCF-QNRF and NWPU: TCFormer reduces the MAE gap to heavy models while remaining far more compact. These results indicate that density-guided aggregation allows our 5M-parameter transformer to encode crowd features nearly as well as much larger networks.

\begin{table}[h]
	\centering
	\caption{Efficiency comparison of lightweight crowd-counting models evaluated on the Kunpeng 920. \textbf{FLOPs} denote the total giga floating-point operations, and \textbf{CPU} latency reflects the single-image inference time. All metrics are measured using a $224 \times 224$ input image.}
	\label{tab:efficiency}
	\begin{tabular}{l|c|c|c}
		\toprule
		\textbf{Method} & \textbf{Params (M)} & \textbf{FLOPs (G)} & \textbf{CPU (ms)} \\
		\midrule
		DHMoE & 5.86 & 9.45 & 2249.13 \\
		RepMobileNet & 3.41 & 2.99 & 736.17 \\ 
		TinyCount & 0.06 & 0.16 & 23.91 \\ 
		\midrule
		\textbf{TCFormer (Ours)} & \textbf{5.52} & \textbf{1.18} & \textbf{329.60} \\
		\bottomrule
	\end{tabular}
\end{table}

\subsection{Efficiency Analysis}
\label{subsec{efficiency}}
To assess the computational efficiency of TCFormer, this paper compares it against three representative lightweight crowd counting models: RepMobileNet, TinyCount, and DHMoE. These methods cover diverse architectural paradigms including re-parameterizable CNN, specialized lightweight CNN, and distillation-based network. To ensure a rigorous and fair comparison, all models are evaluated on the same hardware platform (Kunpeng 920 CPU) under a batch size of 1 with an input resolution of $224 \times 224$.

As summarized in Table~\ref{tab:efficiency}, the TCFormer maintains a highly compact footprint with only 5.52M parameters. While the FLOPs of our model are slightly higher than those of TinyCount and RepMobileNet, it achieves a highly competitive inference latency on the target CPU. Crucially, this marginal increase in computational cost is justified by the significant gains in predictive accuracy across all benchmarks. These findings indicate that the LDWA module and the classification head introduce negligible overhead while substantially enhancing the model’s predictive accuracy. Thus, the results demonstrate that the proposed method offers a superior accuracy-efficiency trade-off, making it well-suited for deployment in resource-constrained real-world environments.

%Table~\ref{tab:efficiency} reports the number of parameters, FLOPs, and measured inference latency. TCFormer contains only 5M parameters while maintaining moderate computational cost. Despite its slightly higher FLOPs relative to TinyCount and RepMobileNet, TCFormer achieves significantly better accuracy (as shown in Sec.~\ref{sec:results}) with only a minimal increase in latency. This indicates that the proposed LDWA module and auxiliary classification head introduce negligible overhead while substantially enhancing predictive performance.

\begin{table*}[tp]
	\centering
	\caption{Ablation study of TCFormer components on the Shanghai Part A dataset. ``DC head" denotes Density-level Classification Head. ``LDWA" represents Learnable Density-Weighted Averaging module}
	\label{tab:ablation_main}
	\begin{tabular}{lcccccc}
		\hline
		\textbf{Model} & \textbf{DC Head} & \textbf{LDWA module} & \textbf{MAE$\downarrow$} & \textbf{RMSE$\downarrow$} & \textbf{Param (K) $\uparrow$} & \textbf{CPU (ms) $\uparrow$} \\
		\hline
		\multirow{3}{*}{\textbf{Baseline}} & $\times$ & $\times$ &70.645 & 117.971 & 5113 & 301.40 \\
		& $\checkmark$ & $\times$ & -5.108& -8.722 & +411.018 & +26.4 \\
		& $\checkmark$ & $\checkmark$ & -6.104 & -7.919 & +411.338 & +28.2 \\
		\hline
	\end{tabular}
\end{table*}

\begin{figure}[!t]
	\centering
	% --- Subfigure (a)
	\subfigure[Efficiency Comparison (CPUs vs Params)]{
			\includegraphics[width=0.30\textwidth]{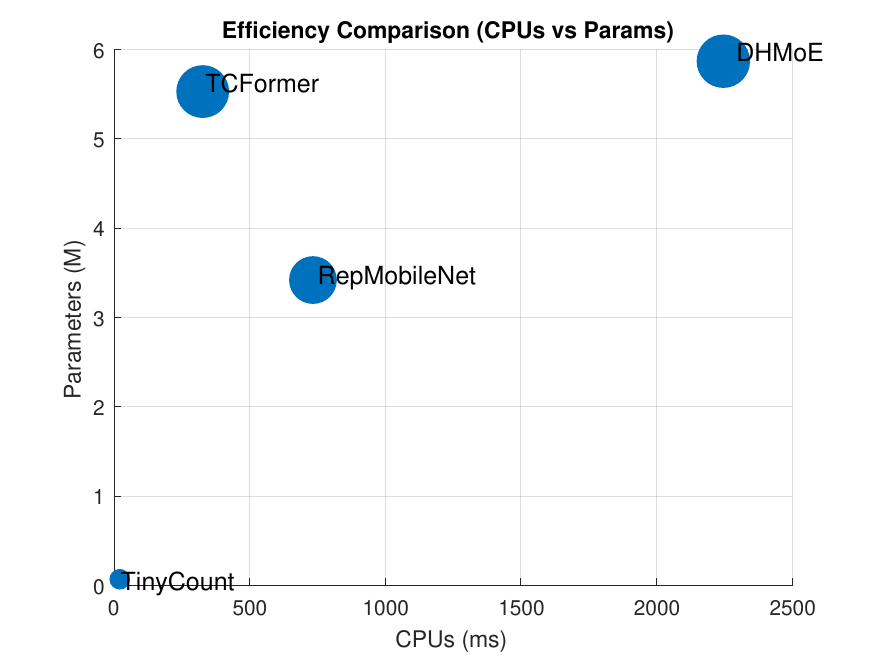}
		}\hfill
	% --- Subfigure (b)
	\subfigure[Prediction--Accuracy Comparison (nMAE and nRMSE)]{
			\includegraphics[width=0.30\textwidth]{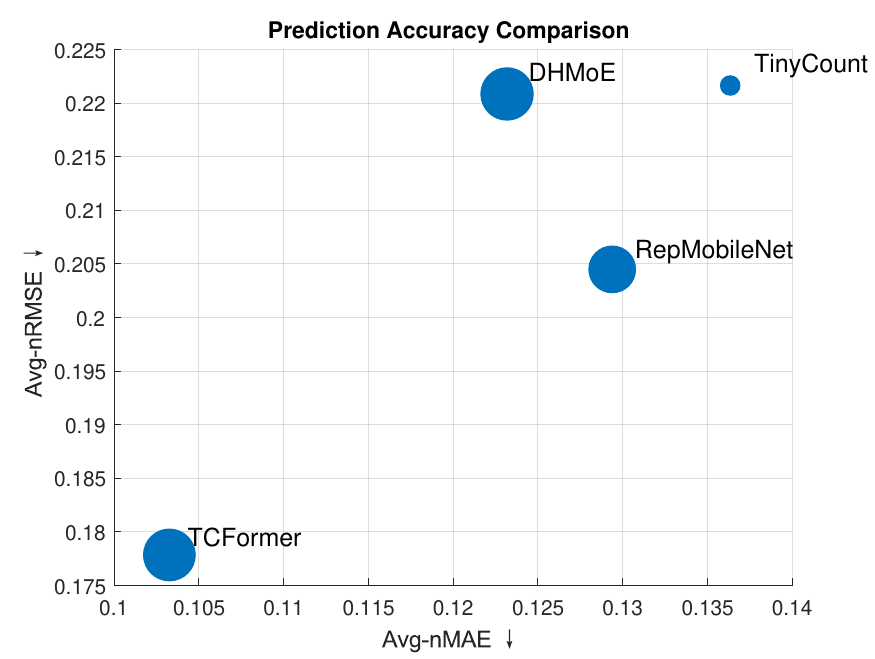}
		}\hfill
	% --- Subfigure (c)
	\subfigure[Efficiency--Accuracy tradeoff]{
			\includegraphics[width=0.30\textwidth]{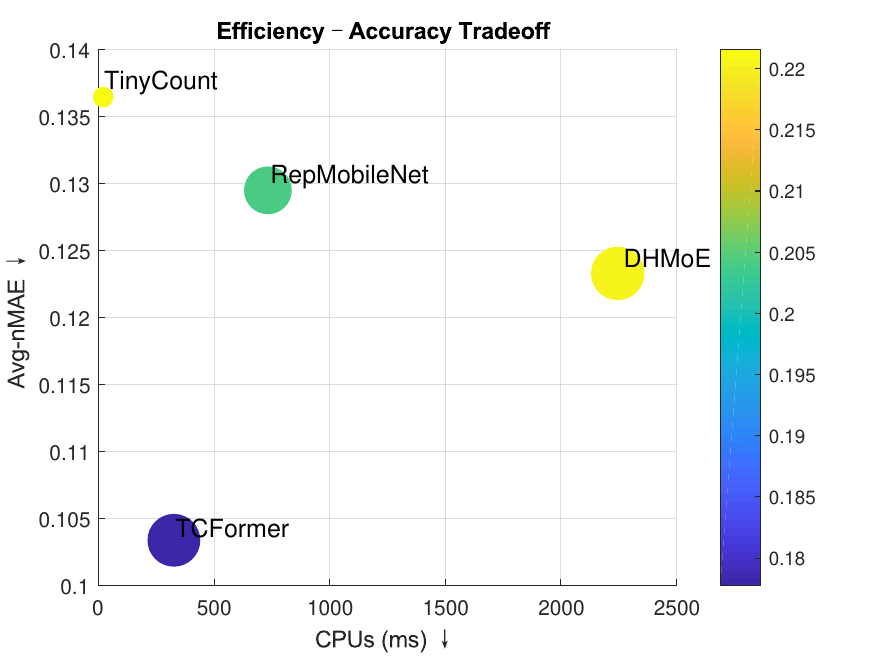}
		}
	
	\caption{
			Comprehensive multi-view comparison of crowd-counting models.  
			(a) FLOPs–Parameters plot showing computational cost.  
			(b) Accuracy comparison using normalized MAE (nMAE) and normalized RMSE (nRMSE).  
			(c) Joint efficiency–accuracy tradeoff, with FLOPs on the x-axis, nMAE on the y-axis, marker size proportional to model parameters, and marker color encoding nRMSE.
		}
	\label{fig:three_views}
\end{figure}

To conduct a multidimensional evaluation of computational efficiency and prediction accuracy, this paper presents a comprehensive assessment from three complementary perspectives in Figures~\ref{fig:three_views}(a)–(c). Each model is represented by a circle whose size is proportional to its total number of parameters. Figure~\ref{fig:three_views}(a) first reveals the relationship between FLOPs operations and model parameters. By comparing with existing paradigms, it characterizes the architectural footprint and raw computational complexity of our framework. Figure~\ref{fig:three_views}(b) aggregates counting performance using nMAE and nRMSE, offering a stabilized comparison of accuracy across diverse benchmarks by scaling error metrics against the dataset’s mean density. Finally, Figure~\ref{fig:three_views}(c) synthesizes the trade-off between efficiency and accuracy, a critical bottleneck in current research, heavyweight models typically lead in accuracy but pose significant deployment challenges due to high computational costs. In this integrated view, FLOPs and nMAE are mapped to axes, with nRMSE values represented via color coding. As evidenced across all three views, the proposed method consistently occupies the southwest locations, demonstrating superior overall efficiency by achieving better accuracy with significantly lower computational costs than representative lightweight models.

%To comprehensively evaluate both computational efficiency and prediction accuracy, 
%we integrate all visualizations into a single figure comprising three subfigures, 
%as shown in Fig.~\ref{fig:three_views}(a)--(c).  
%Fig.~\ref{fig:three_views}(a) presents the relationship between FLOPs and model parameters, 
%reflecting the raw computational cost. Fig.~\ref{fig:three_views}(b) summarizes counting performance using the normalized MAE (nMAE) 
%and normalized RMSE (nRMSE). Fig.~\ref{fig:three_views}(c) provides an integrated efficiency--accuracy tradeoff view, where FLOPs are displayed on the x-axis, nMAE on the y-axis, marker size is proportional to the model parameters, and marker color encodes nRMSE. Together, these three complementary subfigures reveal how each model balances computational cost and predictive accuracy. TCFormer consistently lies near the Pareto-optimal frontier across all views, demonstrating superior overall efficiency.

\subsection{Ablation Studies}
\label{subsec:ablation}
To thoroughly assess the contribution of each component in the proposed TCFormer framework, this paper conducts a series of ablation experiments on the ShanghaiTech Part A dataset. Our study focuses on three progressively enhanced model variants: (1) the baseline TCFormer without extra modules, (2) TCFormer equipped with the density-level classification head and classification loss, and (3) the powerful TCFormer incorporating both the LDWA module and the density-level classification mechanism.

\subsubsection{Effect of Density-Level Classification}
This paper first evaluate the impact of introducing the density-level classification head, which provides additional global to categorize regions of the image based on their congestion levels. By supervising the model with a classification loss, the TCFormer develops a more robust feature representation that is sensitive to varying crowd distributions. As shown in the results, adding this module consistently reduces both 5.108 (MAE) and 8.722 (RMSE), and notably improving performance over the baseline. This demonstrates that even under weak supervision, coarse density categorization helps guide feature learning, and enhances the model's ability to adapt to complex crowd distributions.

\subsubsection{Effect of the LDWA Module}
The LDWA module is designed to refine the final count by dynamically weighing the contributions of different features based on their predicted density. When integrated into the TCFormer, the LDWA module further minimizes estimation errors by 0.996 (MAE), While the improvement in MAE is substantial, the RMSE remained relatively higher at 0.805. This suggests that while the LDWA module effectively reduces the average estimation error across the dataset, it still faces challenges in high-variance scenarios, where outlier errors can disproportionately influence the RMSE metric. 
%Next, we analyze the contribution of the LDWA module, which adaptively re-weights spatial tokens based on predicted density responses. When combined with the density-level classification head, LDWA yields the largest performance gain. The substantial improvement indicates that emphasizing high-density regions during feature aggregation effectively mitigates the absence of pixel-level annotations. LDWA enables TCFormer to form implicit spatial attention maps, focusing the representation on informative crowd regions crucial for accurate count inference.

\subsubsection{Summary of Ablation Findings}
These ablation studies demonstrate that both introduced components contribute significantly to the performance of TCFormer. The density-level classification head injects global scene-level supervisory signals, while the LDWA module strengthens spatial feature aggregation with density-aware weighting. Their combination promote the prediction performance of TCFormer under weak supervision, yielding the best accuracy in benchmark datasets.

\subsection{GPU Power Consumption Analysis}
\label{subsec:power}
To further assess the practicality of the TCFormer in real-world scenarios, this paper analyzes the GPU power consumption of TCFormer and some competitors during inference phase. This paper measures the instantaneous and average power consumption using NVIDIA's \texttt{nvidia-smi} monitoring tool on an A100 GPU. The sampling rate is 0.001s.
%Since TCFormer is designed as an ultra-lightweight architecture with only 5.52 million parameters and low computational overhead (1.18 GFLOPs), it inherently requires less energy compared to conventional fully-supervised large CNNs.

\paragraph{Power Comparison.}
On the A100, TCFormer exhibits an average dynamic power consumption of approximately 59.33\,W during inference, significantly lower than lightweight alternatives such as RepMobileNet, whose power draw typically ranges from 68\,W to 74\,W on the same hardware. This reduction is primarily attributed to TCFormer’s compact feature extractor and efficient LDWA module, which avoid computationally expensive dense attention operations common in other models.
\begin{table}[]
	\begin{center}
		\setlength{\tabcolsep}{0.2em}
		\caption{The energy consumption of all networks}\label{Tbl8}
		\begin{tabular}{llll}
			\hline
			\multirow{2}{*}{\textbf{Models}} & \multicolumn{3}{c}{\textbf{Energy Consumption}} \\ \cline{2-4}
			& Max(W)          & Min(W)            & Ave(W)          \\ \hline
			DHMoE & 61.00 & 57.00 & 57.82 \\
			RepMobileNet & 74.00 & 68.00 & 71.52 \\ 
			TinyCount & 59.00 & 57.00 & 57.17 \\
			\midrule
			\textbf{TCFormer (Ours)} & \textbf{62.00} & \textbf{57.00} & \textbf{59.33} \\ \hline
		\end{tabular}
	\end{center}
\end{table}

\paragraph{Energy Efficiency.}
Using the measured inference latency of 14.87\,ms, TCFormer’s energy cost per image is:
\begin{equation}
	E_{\text{inf}} \approx 59.33\,\text{W} \times 0.0149\,\text{s} = 0.88\,\text{J/image}.
\end{equation}
This places TCFormer among the most energy-efficient crowd counting models, The low GPU power footprint of TCFormer has important implications for deployment in edge-oriented environments. Its energy-efficient design reduces operational costs, increases inference density per GPU, and enables longer operational endurance in power-limited platforms such as UAVs, embedded systems, and smart city sensing devices. These advantages highlight TCFormer’s suitability for scalable, real-time crowd analytics in practical applications.

\begin{figure}[h]
	\centering
	\includegraphics[width=0.4\textwidth]{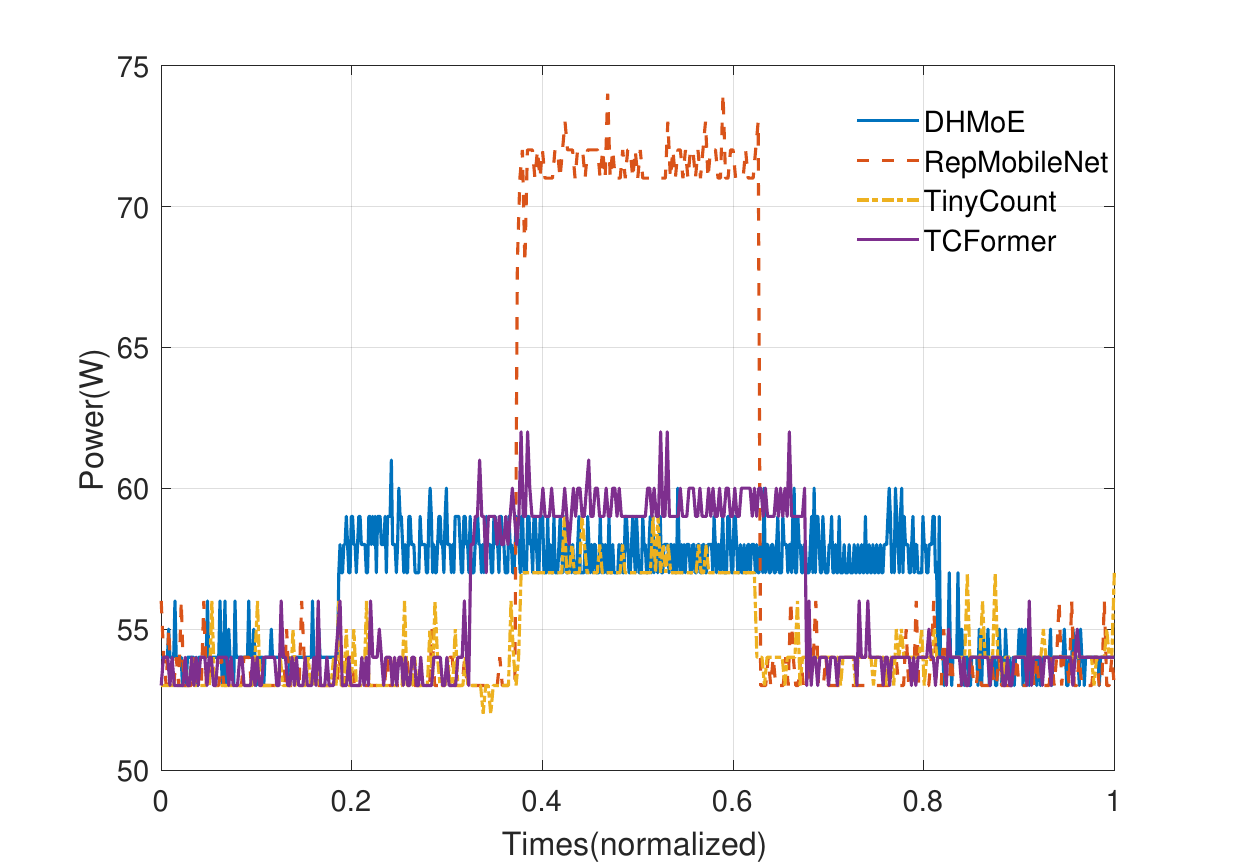}
	\caption{The power consumption curves of the models}
	\label{FIG:pw}
\end{figure}

Fig.~\ref{FIG:pw} draws the power consumption curves of the models, which records the the static idle power and dramatic model's inference power of GPU.

\subsection{Feature visualization Results}
\label{subsec:qualitative}
\begin{figure}[h]
	\centering
	\includegraphics[width=0.4\textwidth]{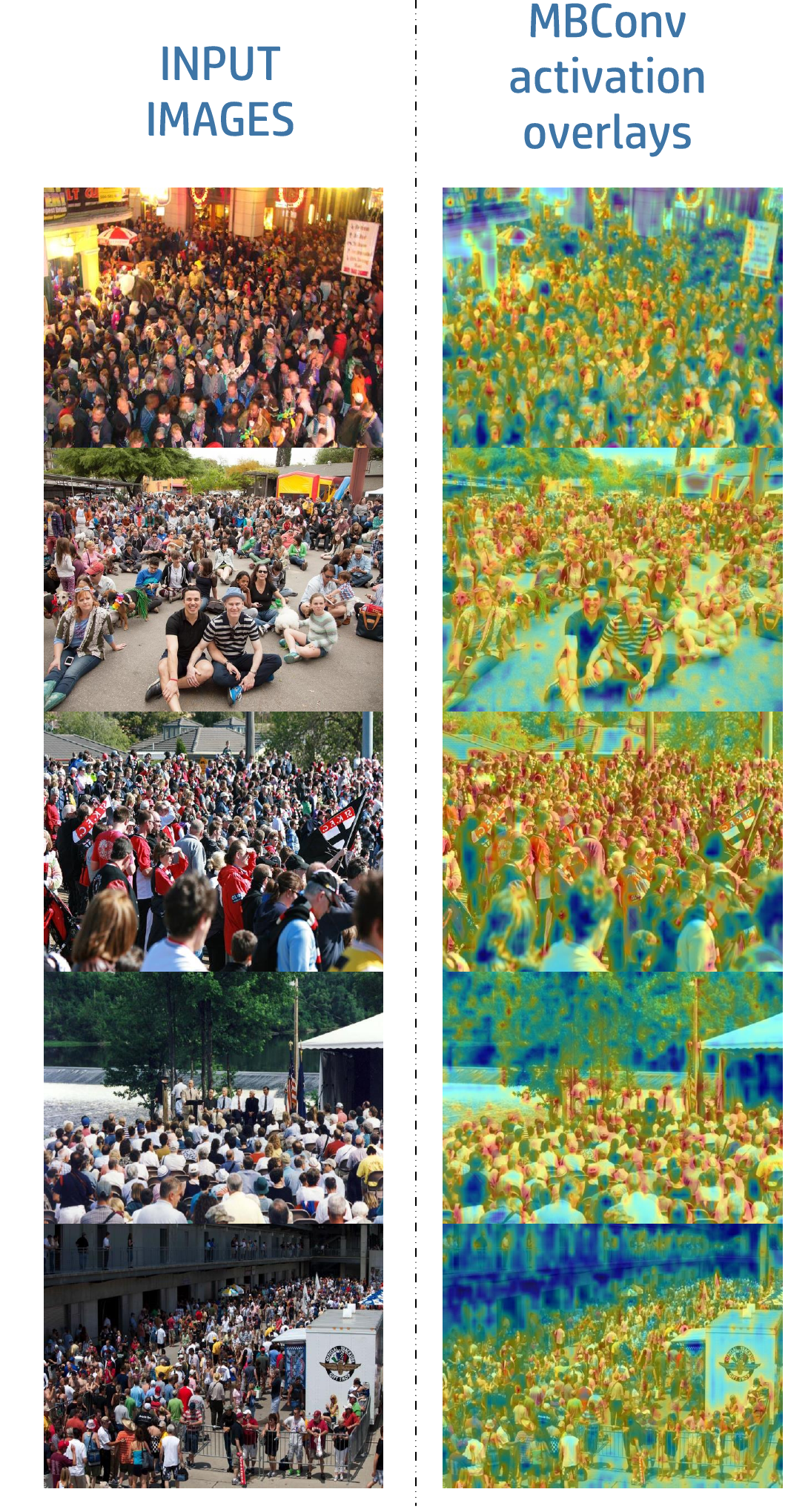}
	\caption{Visualization of Intermediate MBConv Feature Map}
	\label{FIG:mbconv_vis}
\end{figure}

To better understand the internal representations learned by the proposed model, this paper visualizes the intermediate feature maps generated by the MBConv layers within the TinyViT architecture, where both local texture cues and emerging semantic patterns are jointly encoded. As shown in Fig.~\ref{FIG:mbconv_vis}, some representative samples from the ShanghaiTech A dataset have been selected, including scenes with varying illumination and density levels.

Fig.~\ref{FIG:mbconv_vis} illustrates representative MBConv activation maps overlaid on the input images. Despite being learned without any explicit localization supervision,
the MBConv block responses consistently highlight regions with dense crowd presence,
while suppressing background areas such as buildings and open spaces.
This indicates that the convolutional inductive bias embedded in the MBConv blocks
effectively preserves fine-grained local structures that are highly relevant to
crowd density estimation.

\subsection{Discussion}
\label{subsec:discussion}
The experimental results provide several valuable insights into the effectiveness of TCFormer and the underlying design principles of weakly-supervised crowd counting.

\emph{Motivations}: First, the results highlight the importance of global contextual modeling in dense scenes. The TinyViT backbone, though extremely compact, maintains strong long-range dependency modeling capabilities. This global processing complements the density-weighted local aggregation in LDWA, showing that robust crowd understanding requires both global semantics and local density sensitivity.

Second, the experiments demonstrate that introducing spatial inductive bias is essential for weakly-supervised counting. Since only global head counts are available, the model lacks explicit spatial supervision. The proposed LDWA module effectively compensates for this limitation by dynamically re-weighting spatial tokens according to estimated density cues, enabling the network to emphasize congested areas while suppressing backgrounds. This mechanism proves crucial for achieving competitive accuracy without pixel-level annotations. 

Third, the aided density-level classification head significantly stabilizes training. By discretizing the crowd range into density categories, the model gains an additional structured supervisory signal that guides feature learning. This mitigates the ambiguity inherent in weak supervision—particularly for high-density images—and leads to more robust regression performance across a wide range of congestion levels.

Finally, the accuracy efficiency analysis clearly shows that TCFormer achieves a superior balance between model size and performance. Despite using only 5M parameters, TCFormer approaches or exceeds the accuracy of many heavyweight fully-supervised models and consistently outperforms lightweight baselines. This demonstrates the effectiveness of the proposed architecture in real-world, resource-constrained deployment scenarios.

\emph{Methodological Contributions:}
(1) Learnable Density-Weighted Averaging.
LDWA provides an efficient mechanism for spatially adaptive feature aggregation. Unlike global average pooling, which treats all locations equally, LDWA offers a learnable density prior that allows the model to implicitly perceptive crowd distribution. This design is particularly powerful under weak supervision, where explicit density maps are unavailable.

(2) Weakly-Supervised Dual-Head Learning. 
By combining count regression with density-level classification, TCFormer benefits from a multi-task learning framework that strengthens global representation learning. This dual-head design not only enhances the model’s quantitative accuracy but also improves interpretability, as the predicted density level correlates with the model’s internal estimation of scene difficulty.

Overall, the experimental evaluation confirms that TCFormer effectively bridges the gap between \emph{weak supervision}, \emph{efficiency}, and \emph{accuracy}. The proposed framework demonstrates that with carefully designed inductive mechanisms, such as LDWA and density-level classification, a tiny Transformer can achieve performance rivaling large fully-supervised modele requiring neither dense annotations nor extensive computational resources.

%\section{Conclusion}
%
%In this work, we presented \textbf{TCFormer}, an ultra-lightweight Transformer framework for weakly-supervised crowd counting that simultaneously tackles the challenges of annotation cost, computational efficiency, and prediction accuracy. By integrating a compact TinyViT backbone, a learnable density-weighted averaging (LDWA) module, and an auxiliary density-level classification head, TCFormer effectively captures both global context and density-aware spatial cues despite relying solely on image-level supervision. Extensive evaluations on five benchmarks—ShanghaiTech Part A/B, UCF-QNRF, JHU-CROWD++, and NWPU-Crowd—demonstrate that TCFormer, with only 5.1M parameters and 45.2 GFLOPs,  These results verify that carefully designed weakly-supervised strategies and density-aware aggregation can enable tiny models to rival or exceed heavyweight fully-supervised counterparts. The insights gained from this work suggest promising directions for future research, including broader applications of density-aware counting, adaptive density-level modeling, and temporal extensions for video-based crowd analysis. TCFormer sets a new benchmark for efficient, accurate, and annotation-friendly crowd counting, paving the way for practical deployment in real-world, resource-constrained environments.

\section{Conclusion}\label{sec5}
This paper has presented TCFormer, an ultra-lightweight Transformer framework that achieves state-of-the-art crowd counting under weak supervision by integrating three key innovations: a TinyViT backbone for efficient feature extraction, learnable density-weighted averaging for adaptive region focus, and density-level classification for enhanced discrimination. TCFormer effectively captures both global context and density-aware spatial cues despite relying solely on image-level supervision. Extensive experiments demonstrate TCFormer's superiority, outperforming existing methods by 7.4\% average improvement across five benchmarks with only 5.1M parameters and 1.18 GFLOPs. These results verify that carefully designed weakly-supervised strategies and density-aware aggregation can enable tiny models to rival or exceed heavyweight fully-supervised counterparts. Our work establishes that superior accuracy can be achieved without computational burden or dense annotations, providing a practical solution for real-world deployment and opening new avenues for efficient weakly-supervised dense prediction tasks.
\newline

\noindent\textbf{Author Contributions} Qiang Guo: Writing-Original draft preparation, Carrying out the experiments, Methodology, Data curation. Rubo Zhang: Supervision, Reviewing and Editing. Bingbing Zhang ,Junjie Liu, and Jianqing Liu: Contributing to experiments.
\newline

\noindent\textbf{Funding} No funding was received to carry out this study.
\newline

\noindent\textbf{Data availability and access} The benchmark datasets used in this paper come from these papers \cite{zhang2016single,idrees2018,wang2020nwpu}. %The SY-Metro dataset are not available due to commercial restrictions.

\section*{Declarations}
\noindent\textbf{Competing Interests} The authors have no competing interests to disclose in any material discussed in this article.
\newline

\noindent\textbf{Ethical and informed consent for data used} Not applicable.
\newline

\end{document}